\documentclass[a4paper,10pt]{article}
\usepackage{amsmath,amsfonts,amssymb,amsthm}
\usepackage{graphicx,subfigure,booktabs,multirow}
\usepackage{mathrsfs,enumerate,color}
\usepackage{algorithm}
\usepackage[textwidth=14cm]{geometry}
\usepackage[sort&compress, square, numbers]{natbib}
\usepackage{hyperref}

\usepackage{diagbox}
\usepackage{threeparttable}
\usepackage{booktabs}

\usepackage{latexsym,bm, amsthm}
\usepackage[normalem]{ulem}
\usepackage{tabularray}

\useunder{\uline}{\ul}{}

\usepackage{algorithmic}
\usepackage{graphicx}

\newtheorem{theorem}{Theorem}[section]

\newtheorem{definition}{Definition}[section]

\newcommand{\iq}{{\bf i}}
\newcommand{\jq}{{\bf j}}
\newcommand{\kq}{{\bf k}}

\newcommand{\pq}{{\bf p}}

\newcommand{\qq}{{\bf q}}

\newcommand{\Aq}{{\bf A}}
\newcommand{\aq}{{\bf a}}

\newcommand{\Cq}{{\bf C}}

\newcommand{\vq}{{\bf v}}

\newcommand{\Gq}{{\bf G}}

\newcommand{\xq}{{\bf x}}

\newcommand{\zq}{{\bf z}}

\newcommand{\yq}{{\bf y}}
\newcommand{\Wq}{{\bf W}}
\newcommand{\wq}{{\bf w}}

\newcommand{\Qq}{\mathbb{Q}}

\title{Quaternion  Generative Adversarial Neural Networks \\
and Applications to Color Image Inpainting
}

\author{
Duan Wang\thanks{Duan Wang was with School of Mathematics and Statistics, Jiangsu Normal University,
Xuzhou 221116, P. R. China. E-mail: duanw@jsnu.edu.cn},
        Dandan Zhu\thanks{Dandan zhu was with School of Mathematics and Statistics, Jiangsu Normal University,
	Xuzhou 221116, P. R. China. E-mail: dandanzh186@163.com},
        Meixiang Zhao\thanks{Meixiang Zhao was with School of Mathematics and Statistics, Jiangsu Normal University,
	Xuzhou 221116, P. R. China. E-mail: zhaomeixiang2008@126.com},
        Zhigang Jia\thanks{Corresponding author. Z. Jia was with School of Mathematics and Statistics, Jiangsu Normal University,
	Xuzhou 221116, P. R. China. E-mail: zhgjia@jsnu.edu.cn}
}

\begin{document}
\date{}
\maketitle
\begin{abstract}
Color image inpainting is a challenging task in imaging science. The existing method is based on real operation, and the red, green and blue channels of the color image are processed separately, ignoring the correlation between each channel. In order to make full use of the correlation between each channel, this paper proposes a Quaternion Generative Adversarial Neural Network (QGAN) model and related theory, and applies it to solve the problem of color image inpainting with large area missing. Firstly, the definition of quaternion deconvolution is given and the quaternion batch normalization is proposed. Secondly, the above two innovative modules are applied to generate adversarial networks to improve stability. Finally, QGAN is applied to color image inpainting and compared with other state-of-the-art algorithms. The experimental results show that QGAN has superiority in color image inpainting with large area missing.\\

\end{abstract}

\section{Introduction}\label{sec1}
In 1843, the concept of quaternions was first introduced by Sir William Rowan Hamilton. Nowadays, quaternions have emerged as a crucial mathematical concept in various fields such as Mechanics \cite{m1, m2, m3}, Optics \cite{m4, s45}, Modern Computer Graphics \cite{m5, s35}, Color Image Processing \cite{s34, m10, s33}, and more. As a new tool for color image representation, quaternions have obtained remarkable results in color image processing. The three imaginary parts of the quaternion correspond to the three channels of the color image, and each color image corresponds to a quaternion matrix, and the quaternion can achieve the fusion of color information by processing the color information of different color channels simultaneously. Therefore, quaternion is applied to convolutional neural networks. 

The pioneering work of Zhu et al. introduces the Quaternion Convolutional Neural Networks (QCNN) \cite{s1}, which incorporates network layers such as convolutional layer and fully connected layer. In a similar vein, Parcollet et al. propose the Quaternion Recurrent Neural Networks (QRNN) \cite{s2} as an extension to traditional real-valued RNN. QRNN significantly reduces the number of free parameters required, thereby enabling a more compact representation of the association information.

Color image inpainting is a challenging task in the field of image processing. The objective of completion is to reconstruct areas of an image that are either missing or damaged. It has many applications, such as repairing damaged paintings or photographs \cite{s3}. Classical image inpainting methods are usually based on local or non-local information to inpaint an image. Local methods rely on the presence of priori information in the input image. For example, missing or corrupted parts of a texture image can be inpainted by finding the nearest patch from the same image \cite{s4} . The full variational approach \cite{s5} takes into account the smoothness of natural images, thus inpainting in small regions of missing and removing pseudo-noise. Some subsets of images may also contain special properties, such as planar structures \cite{s6} or low-rank structures \cite{s7}. But these methods require a priori information. To solve the problem of inpainting images with large area missing, non-local methods try to use external data to predict the missing pixels. Hays et al. \cite{s8} proposed a feasible method to search for semantically similar patches to inpaint in the damaged area from a large library of images. Internet search \cite{s9} can also be used to inpaint the target region of a scene. Both methods require exact matching from databases or the Internet, and are prone to failure when the test scene differs significantly from any database image. 

Unlike previous manual matching and editing, learning-based methods have yielded the desired results \cite{s10,s11,s12,s13}. In 2016,  Pathak et al. proposed an unsupervised visual feature learning algorithm based on contextual pixel prediction \cite{s14}. In 2017, Yeh et al. proposed a new semantic image inpainting method \cite{s15}, which treats image inpainting as an image generation problem with constraints. By introducing convolution and generative adversarial networks(GAN), it can better inpaint images with large area missing. In 2021, Liu et al. proposed probabilistic diverse GAN for images inpainting \cite{s36}, which can create many inpainting outputs withdiverse and visually realistic material. In 2022, Zheng et al. propose cascaded modulation GAN consisting of an encoder with Fourier convolution blocks that extract multi-scale feature representations from the image \cite{s37}. These designs increase the ability of GAN image inpainting. However, the above image inpainting algorithms are all based on real number operations, which operate separately for the three channels of the color image, and the correlation between the channels is not guaranteed.

As a new color image representation tool, quaternion is used to process the color information of different color channels simultaneously by representing the color image as a quaternion matrix to achieve the fusion of color information. In 2019, Z. Jia et al. proposed the robust quaternion matrix completion (QMC) algorithm \cite{s23} and applied it to color image inpainting, in which the robust color image completion can be solved by a convex optimization problem in the quaternion framework. In 2020, Miao et al. introduced a novel approach, termed the low-rank quaternion tensor complementation algorithm\cite{s16}, for the restoration of color video images. This algorithm employs the alternating direction method of multipliers (ADMM) framework to optimize the model, guaranteeing algorithm convergence, and delivering an arbitrary-order quaternion tensor complementation algorithm. In 2022, J. Miao et al. proposed a new low-rank quaternion matrix completion algorithm \cite{s17} to recover missing data from color images. The two methods, low-rank decomposition and kernel norm minimization, were combined to solve the model based on quaternion matrices using the alternating minimization method. In 2022, Z. Jia et al. proposed a new NSS-based QMC algorithm using Nonlocal Self-Similarity (NSS) \cite{s24}, and applied it to solve the color image inpainting problem; the NSS-based QMC algorithm computes the best low-rank approximation to obtain a higher quality color inpainted image. A new QMC method based on the quaternion tensor NSS is proposed to solve the color video inpainting problem. However, the above quaternion color image inpainting methods still use local information to inpaint color images, which may not achieve the desired effect for inpainting images with large area missing. 

So far, the research on inpainting color images with large area missing based on quaternions needs to be further improved. Despite the concept of Quaternion  Generative Adversarial Neural Networks having been introduced\cite{r1} \cite{r2}, its theoretical underpinnings still lack depth, and the scope of its application remains overly narrow. In this paper, we propose a new generative adversarial neural networks based on quaternions, and apply it to color image inpainting.

\begin{figure}[htbp]
	\centering
	\includegraphics[width=0.95\textwidth,height=0.5\textwidth]{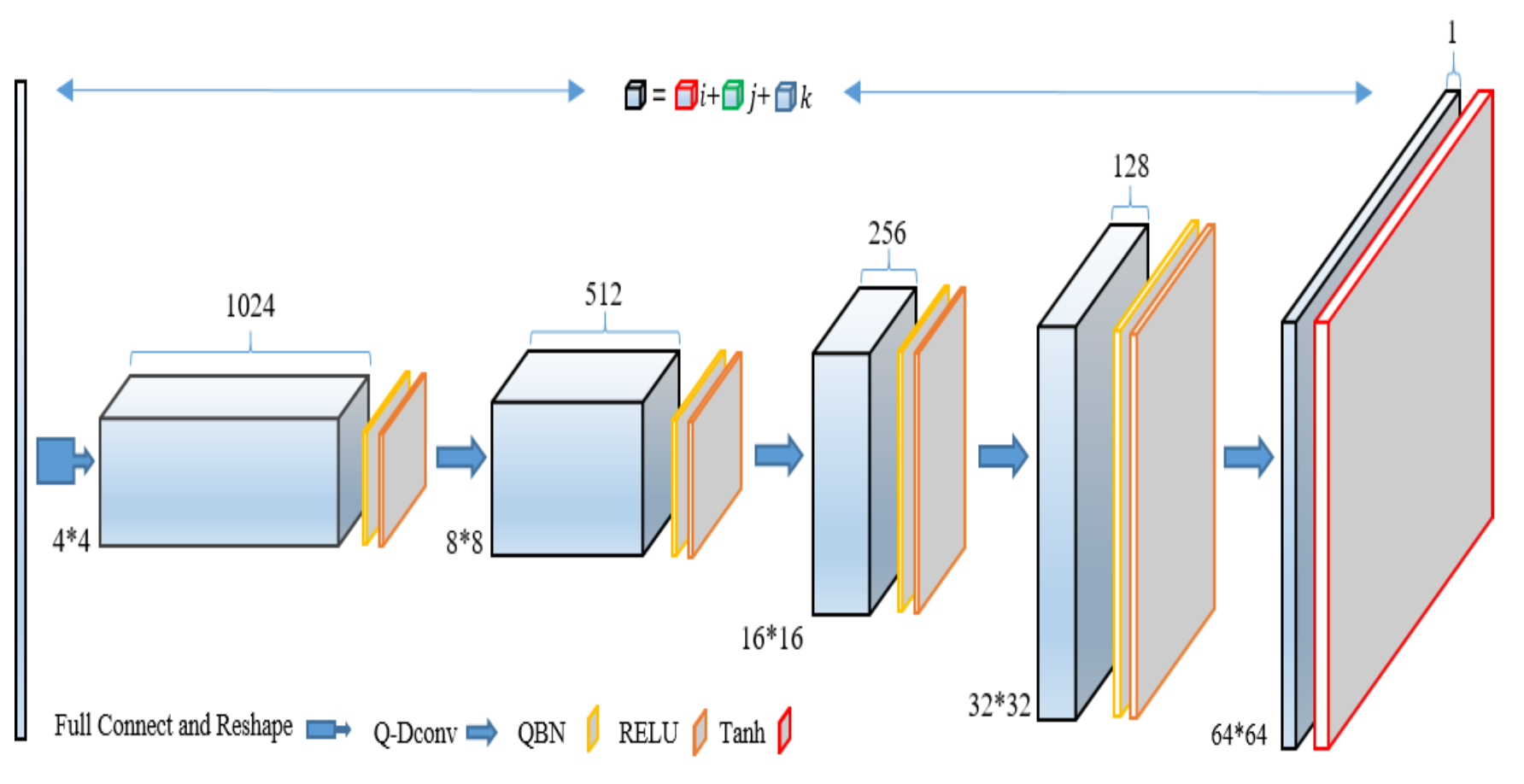}

	\caption{Schematic of a quaternion generative adversarial network.}
	\label{f:C-G}
\end{figure}

\section{Related Work}\label{sec:background}
\textbf{Quaternion.}
Initially, the definition of quaternion and its often utilized associated notions are presented. The set of real numbers is denoted by $\mathbb{R}$.  The set of quaternions is denoted by $\mathbb{Q}$, where the elements are often denoted as $\qq$.
\begin{definition}
	Any quaternion $\qq \in \mathbb{Q} $ is denoted as \cite{s19}
	$$
	\qq=q_{0}+q_{1}\iq+q_{2}\jq+q_{3}\kq,
	$$
	where $q_{0}, q_{1}, q_{2}, q_{3}$ are real numbers,  $\iq, ~\jq,~ \kq$ are three imaginary parts. And the three imaginary parts have the following relations.
	$$
	\iq^{2}=\jq^{2}=\kq^{2}=\iq\jq\kq=-1.
	$$
	If $q_{0}=0$, then the quaternion is said to be a pure imaginary quaternion.
\end{definition}

Quaternions can also be represented by scalars and vectors \cite{s21}:
$${\bf q}=s+ \vec{\vq},$$
where $s$ means that the scalar part corresponds to the $q_{0}$ component of $\qq$, and $\vec{v}$ means that the vector part corresponds to the $i, j, k $ component of $\qq$. A pure imaginary quaternion can be directly represented by a 3D vector. For convenience, in this paper, for any quaternion \qq, its corresponding vector form is denoted as $\vec{\qq}$.

The conjugate of the quaternion $\qq=q_{0}+q_{1}\iq+q_{2}\jq+q_{3}\kq$ \cite{s19} is denoted by $\bar{\qq}$:
$$\bar{\qq}=q_{0}-q_{1}\iq-q_{2}\jq-q_{3}\kq=s-\vec{{\bf v}}.$$

The modulus of a quaternion \cite{s34} is noted as $|{\bf q}|$,
$$|{\bf q}|=\sqrt{{\bf q} {\bf q}^{*}}=\sqrt{{\bf q}^{*} {\bf q}} =\sqrt{q_{0}^{2}+q_{1}^{2}+q_{2}^{2}+q_{3}^{2}},$$
The quaternion called $|{\bf q}|=1$ is the unit quaternion.

The inverse  of a nonzero quaternion \cite{s19} is denoted as ${\bf q}^{-1}$.
$${\bf q}^{-1}=\frac{\bar{\qq}}{|{\bf q}|} .$$

The  $n$-dimensional quaternion vector \cite{s29} is denoted as 
$$\Qq=[\qq_{1}, \qq_{2},\ldots, \qq_{n} ]^{T} \in \mathbb{Q}^{n}.$$

For a number of quaternions ${\bf q}_{l}=w_{l}+x_{l}i+y_{l}j+z_{l}k, l=1, 2, \ldots, n$. The mean and variance of a quaternion are defined as
\begin{equation}\label{e:EV}
	\begin{split}
		E({\bf q})&=\frac{1}{n}\sum \limits_{l=1}^{n}{\bf q}_{l}=E[w]+E[x]i+E[y]j+E[z]k,\\
		& Var({\bf q})=\frac{1}{n}\sum \limits_{l=1}^{n}\|{\bf q}_{l}-E[{\bf q}]\|^{2}.
	\end{split}
\end{equation}

The quaternion matrix $\Aq \in \mathbb{Q}^{m \times n}$ can be written as
$$\Aq=A_{0}+A_{1}\iq+A_{2}\jq+A_{3}\kq,$$
where $ A_{0}, A_{1}, A_{2}, A_{3} \in \mathbb{R}^{m \times n}.$

The $l_{1}$-parametrization of the quaternion matrix $\Aq \in \mathbb{Q}^{m \times n}$ \cite{s23} is 
\begin{equation}\label{e:23}
	\|\Aq\|_{1}=\sum \limits_{i=1}^{m} \sum\limits_{j=1}^{n} |a_{ij}|.
\end{equation}

\textbf{Generating Adversarial Networks.}
Generative Adversarial Networks (GAN) \cite{s25}  is a popular generative modeling framework to generate high-quality natural images. The framework has two network models: Generator, Discriminator. Discriminator is responsible for discerning the authenticity of an image and generator is required to produce images that possess a high degree of realism. Deep convolutional generative adversarial neural network (DCGAN) \cite{s28} incorporate convolutions into the GAN model. A multitude of GAN frameworks have emerged base on this foundation.

Since the development of GAN, its loss function has been updated based on the initial GAN loss function. The initial GAN loss function is 
\begin{equation}\label{e:GAN}
	\begin{split}
		&\mathop{\rm min}\limits_{G} \mathop{\rm max}\limits_{D}V(G,D) = \\
		&E_{x\sim p_{data}}[{\rm log}(D(x))]+E_{z\sim p_{z}}[{\rm log}(1-D(G(z))],
	\end{split}
\end{equation}
where $x$ represents the real image and the random vector $z$ denotes the input of $G$. The loss function corresponding to the discriminator is ${\rm log}(D(x))+{\rm log}(1-D(G(z))$, and the loss function corresponding to the generator is ${\rm log}(D(G(z)). $

\textbf{Quaternion Generative Adversarial Networks.}
While scholars have proposed Quaternion Generative Adversarial Networks \cite{r1} \cite{r2}, their theoretical foundation remains deficient. In paper \cite{r2}, the authors, for the sake of generator stability, adopt the UNET network architecture as the generator model, extending the data input to four dimensions and convolving it. Thus, its essence is still the integration of the three-channel real convolutions of images, neglecting the rotational characteristics of quaternions. Paper \cite{r1} also treats the deconvolution in QGAN as real value operation. This leads to an incomplete QGAN proposed, necessitating a more comprehensive theory to elucidate quaternion deconvolution.

\textbf{Quaternion convolution.}
A new quaternion convolutional neural network (QCNN) model \cite{s1} is proposed for representing colour images in quaternions to address the loss of correlation between channels \cite{m9} in convolutional neural network (CNN) . Each colour pixel in a colour image is represented as a quaternion, so the image is represented as a matrix of quaternions, rather than as three separate matrices of real values. A quaternion convolution layer is designed using the quaternion matrix as the input to the network. Traditional real-valued convolution only scales and transforms the input, whereas quaternion convolution scales and rotates the input in colour space, providing a more structured representation of colour information.

\begin{definition} \cite{s1} (Quaternion convolution)\label{qcnn}
	Suppose a quaternion matrix  $\Aq=[\aq_{ij}] \in \mathbb{Q}^{m \times n} $ represents a colour image. Then we have a quaternion convolution kernel $\Wq=[\wq_{ij}] \in \mathbb{Q}^{d \times d} $. The elements of the convolution kernel are $
	\wq_{ij}=s_{ij}({\rm cos} \frac{\theta_{ij}}{2}+{\rm sin} \frac{\theta_{ij}}{2}\boldsymbol{\mu})$, where $\theta_{ij} \in [-\pi,\pi]$ is the rotation angle, $ s_{ij} \in R $ is the scaling factor, and the rotation axis $\boldsymbol{\mu}$ is the unit-length gray axis (i.e., $\frac{\sqrt{3}}{3}(i +j + k))$. The quaternion convolution of an image A about W is defined as
	\begin{equation}\label{AW}
		A*W=[\zq_{i^{'}j^{'}}] \in \mathbb{Q}^{(m-d+1)\times (n-d+1)}.
	\end{equation}
	where
	\begin{equation}\label{AW1}
		\zq_{i^{'}j^{'}}=\sum\limits_{i=1}^{d}\sum\limits_{j=1}^{d}\frac{1}{s_{ij}}w_{ij}a_{(i^{'}+i) (j^{'}+j)}w_{ij}^{*}.
	\end{equation}
\end{definition}

If each quaternion $\aq_{ij}$ is represented by a three-dimensional column vector, denoted $\vec{a}_{ij}$. Then according to quaternion rotation, the operations in Equation $\eqref{AW1}$ can be expressed as a set of matrix multiplications
\begin{equation}\label{KM}
	\vec{\zq}_{i^{'}j^{'}}=\sum\limits_{i=1}^{d}\sum\limits_{j=1}^{d}s_{ij}
	\left[
	\begin{array}{cccc}
		\gamma_{1} & \gamma_{2}& \gamma_{3} \\
		\gamma_{3} &\gamma_{1} &\gamma_{2} \\
		\gamma_{2} &\gamma_{3} &\gamma_{1}
	\end{array}
	\right]\vec{\aq}_{(i^{'}+i)(j^{'}+j)},
\end{equation}
where $\gamma_{1}=\frac{1}{3}+\frac{2}{3}{\rm cos}\theta_{ij},\gamma_{2}=\frac{1}{3}-\frac{2}{3}{\rm cos}(\theta_{ij}-\frac{\pi}{3}),\gamma_{3}=\frac{1}{3}-\frac{2}{3}{\rm cos}(\theta_{ij}+\frac{\pi}{3})$.

It can be seen that the output $\vec{\zq}_{i^{'}j^{'}}$ is also a quaternion vector. Let the rotation matrix in Equation $\eqref{KM}$ denoted by $R_{ij}$. Equation $\eqref{KM}$ can be written in the following form
\begin{equation}\label{KM1}
	\vec{\zq}_{i^{'}j^{'}}=\sum\limits_{i=1}^{d}\sum\limits_{j=1}^{d}s_{ij}R_{ij}\vec{\aq}_{(i^{'}+i)(j^{'}+j)}.
\end{equation}

\begin{theorem} \cite{s1} \label{g:0989}
	Let $\qq$ be a pure quaternion and $\wq=s({\rm cos} \frac{\theta}{2}+{\rm sin} \frac{\theta}{2}\boldsymbol{\mu})$ be the quaternion convolution kernel element, where $\theta \in [-\pi,\pi]$ is the rotation angle, $ s \in R $ is the scaling factor, and the rotation axis $\boldsymbol{\mu}$ is the unit-length gray axis (i.e., $\frac{\sqrt{3}}{3}(i +j + k))$. The quaternion convolution operation $\pq=\frac{1}{s}\wq \qq\wq^{*}$ corresponds to a matrix of the form $\vec{\pq}=sR\vec{\qq}$, where R is the rotation matrix corresponding to the convolution operation. The partial derivatives of the output variable \pq ~with respect to the input variable \qq ~and the rotation parameters $s, \theta$ are
	$$\frac{\partial {\vec{\bf p}}}{\partial{\vec{\bf q}}}=sR, ~\frac{\partial {\vec{\bf p}}}{\partial s}=R\vec{\qq},  ~\frac{\partial {\vec{\bf p}}}{\partial \theta}=sR'\vec{\qq},$$
	where $\vec{\pq}$, $\vec{\qq}$ are the vector forms corresponding to the quaternions $\pq$, $\qq$ respectively.
\end{theorem}
Then we can know the derivative of the quaternion convolution rule:
\begin{equation}\label{e:qbp}
	\frac{\partial loss}{\partial {\vec{\bf q}}}=\frac{\partial loss }{\partial {\vec{\bf p}}}\frac{\partial{\vec{\bf p}}}{\partial{\vec{\bf q}}}=\frac{\partial loss }{\partial {\vec{\bf p}}}sR.
\end{equation}

\textbf{Semantic inpainting algorithm}

	$\bullet$ {\bf Contextual pixel prediction:} For the color image inpainting algorithm based on contextual pixel prediction in paper \cite{s14}, the reconstruction loss $L_{rec}$ and the adversarial loss $L_{adv}$ are respectively
\begin{equation}\label{GAN}
		L_{rec}(x)=||\hat{M}\odot (x-F((1-\hat{M})\odot x))||_{2}^{2},
\end{equation}
\begin{equation}\label{GAN}
	\begin{split}
		L_{adv}=&\mathop{\rm max}\limits_{D}E_{x\sim p_{data}}[{\rm log}(D(x|y))] \\
		&+E_{z\sim p_{z}}[{\rm log}(1-D(F((1-\hat{M})\odot x))],
	\end{split}
\end{equation}
where the binary mask denoted as $\hat{M}$ a matrix that has the same dimensions as the image. If a pixel is absent at a certain place in the image, the corresponding position in the $M$ matrix is assigned a value of 0. Otherwise, it is assigned a value of 1.

The final loss is
\begin{equation}\label{GAN}
	L=\lambda_{rec}L_{rec}+\lambda_{adv}L_{adv}.
\end{equation}

$\bullet$ {\bf GAN model prediction:} The semantic inpainting algorithm for color images based on GAN \cite{s15} is development of above color image inpainting algorithm. It has the context loss $L_{c}$ and the a priori loss $L_{p}$ as follows:
\begin{equation}\label{GAN}
	\begin{split}
		L_{c}(z)=\|W \cdot (G(z)- y))\|_{1},\\
		L_{p}(z|y, M)=\lambda {\rm log}(1-D( G(z))),
	\end{split}
\end{equation}
The final loss is
$$L=L_{c}(z)+L_{p}(z|y, M).$$

\section{Quaternion Generative Adversarial Network}

This section proposes a new generative adversarial neural network model combining quaternions with generative adversarial neural networks.

\subsection{Quaternion deconvolution}\label{sec:3.2}

According to the relationship between real-valued convolution and deconvolution \cite{s40,s41} which is a special linear transformations, we propose the concept of quaternion deconvolution. Assume the output $\yq \in \mathbb{Q}^{m}$, the input $\xq \in \mathbb{Q}^{n}$, and the sparse matrix $\Cq \in \mathbb{Q}^{m \times n}$, where the non-zero elements are the elements of the convolution kernel. We have  $$\yq_{i}=\sum \limits_{j=1}^{n} \frac{1}{s_{ij}}\wq_{ij}\xq_{j}\wq_{ij}^{*}.$$

Then let ${\hat{R}} = 
{\left[
	\begin{array}{cccccccccccccccc}
		s_{11}R_{11} & s_{12}R_{12}  &\cdots & s_{1n}R_{1n}\\
		s_{21}R_{21} & s_{22}R_{22}  &\cdots & s_{2n}R_{2n}\\
		\vdots & \vdots  &\ddots & \vdots\\
		s_{m1}R_{m1} & s_{m2}R_{m2}  &\cdots & s_{mn}R_{mn}
	\end{array}
	\right]}
,$
 we can get form of matrix product:

 \begin{equation}\label{e:38}
	\yq= \hat{R} \xq
\end{equation}

Assume  $\frac{\partial loss}{\partial y_{i}}$  has been obtained from the deeper network, we can get:

\begin{equation}\label{e:39}
\frac{\partial loss}{\partial x} = \hat{R}^T \frac{\partial loss}{\partial y}
\end{equation} 

Equations \eqref{e:38}, \eqref{e:39} show that, in contrast to feedforward propagation, the input to backpropagation is a m-dimensional quaternion vector and the output is a n-dimensional quaternion vector. And the loss is back-propagated by multiplying transpose of the rotation matrix. This operation is called quaternion deconvolution.

Let quaternion matrix $\Aq=[\aq_{ij}]  \in \mathbb{Q}^{m \times n}$ and quaternion convolution kernel $\Wq=[\wq_{ij}]  \in \mathbb{Q}^{d \times d}$. These elements are similar to Definition $\ref{qcnn}$. The operation called quaternion deconvolution of $\wq_{ij}$ on $\aq_{ij}$ can be written a 
$$
\aq_{ij} \circledast \wq_{ij}=s_{ij}R_{ij}^{T}\vec{\aq}_{ij},
$$
where $\circledast$ represents the deconvolution operation. Comparing with Equation $\eqref{e:qbp}$, we find that the backpropagation operation of convolution is the same as quaternion deconvolution. 

It is important to note that each pixel can be represented by a pure quaternion or a 3D vector within the colour space in quaternion deconvolution. Similar to quaternion convolution, quaternion deconvolution only uses rotation and scaling transformations. And those techniques aims to find the optimal representation within a limited region of the colour space. So the process of convolution and deconvolution applies an implicit regularisation to the model, thus minimising the potential overfitting that might arise from excessive degrees of freedom in kernel learning.
The derivative of quaternion deconvolution is given by Theorem \ref{g:0999}.

\begin{theorem}\label{g:0999}
	Let $\qq$ be a pure quaternion and $\wq=s({\rm cos} \frac{\theta}{2}+{\rm sin} \frac{\theta}{2}\boldsymbol{\mu})$ be the quaternion convolution kernel element, where $\theta \in [-\pi,\pi]$ is the rotation angle, $ s \in R $ is the scaling factor, and the rotation axis $\boldsymbol{\mu}$ is the unit-length gray axis (i.e., $\frac{\sqrt{3}}{3}(i +j + k))$. The quaternion deconvolution operation $\pq= \qq\circledast \wq$ corresponds to a matrix of the form $\vec{\pq}=sR^{T}\vec{\qq}$, where $R^{T}$ is the rotation matrix corresponding to the deconvolution operation. The partial derivatives of the output variable \pq~ with respect to the input variable \qq~ and the rotation parameters $s, \theta$ are
	$$\frac{\partial \vec{{\bf p}}}{\partial \vec{\qq}}=sR^{T}, \frac{\partial \vec{{\bf p}}}{\partial s}=R^{T}\vec{\qq}, \frac{\partial \vec{{ \bf p}}}{\partial \theta}=s(R^{T})'\vec{\qq}. $$
	where $\vec{\pq}$, $\vec{\qq}$ are the vector forms corresponding to the quaternions $\pq$, $\qq$ respectively.
\end{theorem}

\begin{figure}[htb]
	\centering
	\centering
	\includegraphics[width=0.95\textwidth,height=0.5\textwidth]{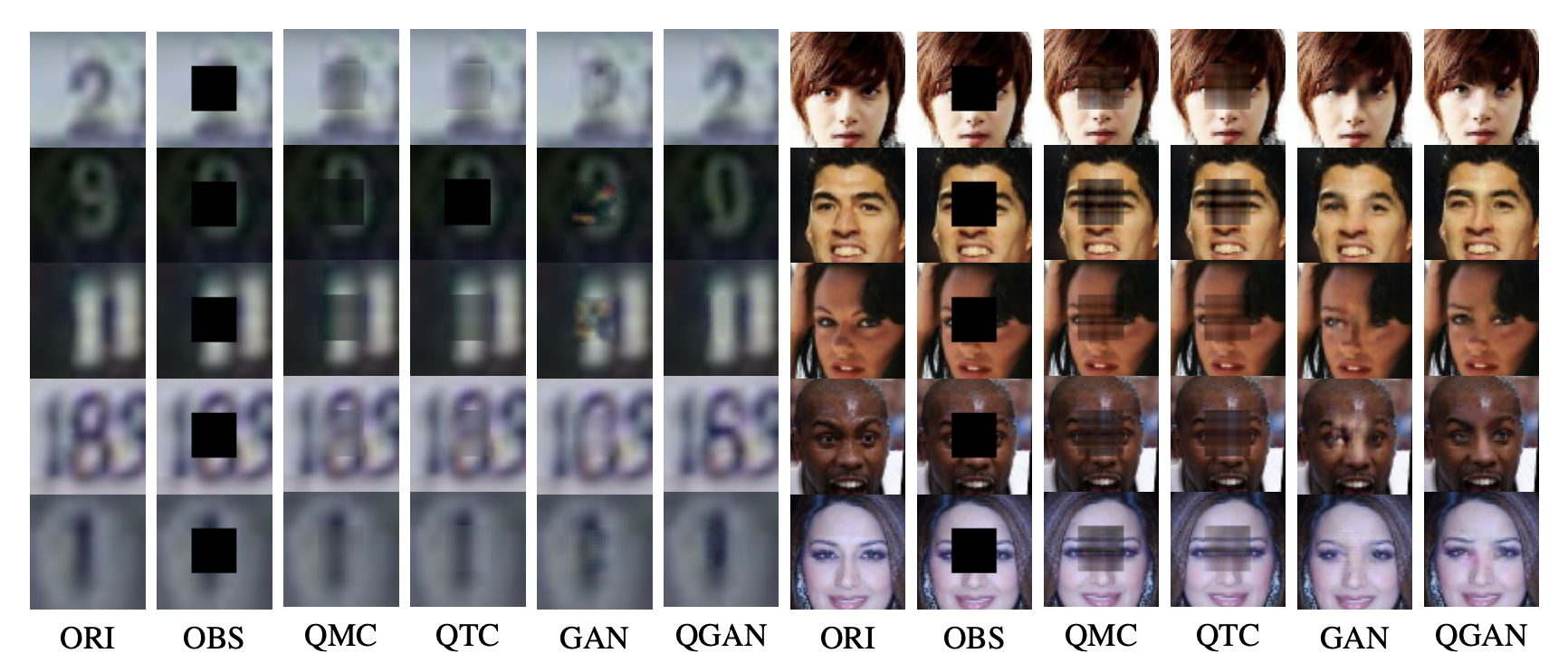}
	
	\caption{SVHN Database $16\%$ center Pixel Deletion Comparison of Various Inpainting Methods.}
	\label{p:SVHN}
\end{figure}

\subsection{Quaternion batch normalization}\label{sec:3.3}

For a layer with d-dimensional inputs $\xq=(\xq^{1}, \xq^{2},\cdots, \xq^{d}) \in \mathbb{Q}^{d}$, we will normalize each dimension
$$
\hat{\xq}^{k}=\frac{\xq^{k}-E[\xq^{k}]}{\sqrt{Var[\xq^{k}]}} , k=1, 2, \ldots, d,
$$
where k represents the k-th dimension of the vector \xq~, ~$E[\xq^{k}]$,~$Var[\xq^{k}]$~ are the expectation and variance corresponding to the k-th dimension.

The above quaternion normalization is one of the simplest quaternion batch normalization, which reduces the effect of ICS to a certain extent \cite{s18}, but limits the expressive power of the data. To solve this problem, the following linear transformations are applied to the normalized data:
$$\yq^{k}=\gamma^{k} \hat{\xq}^{k}+\beta^{k},$$
where $\gamma^{k}$ is a real number and $\beta^{k}$ is a quaternion. 

\begin{figure}[t]
	\centering
	\includegraphics[width=0.8\linewidth]{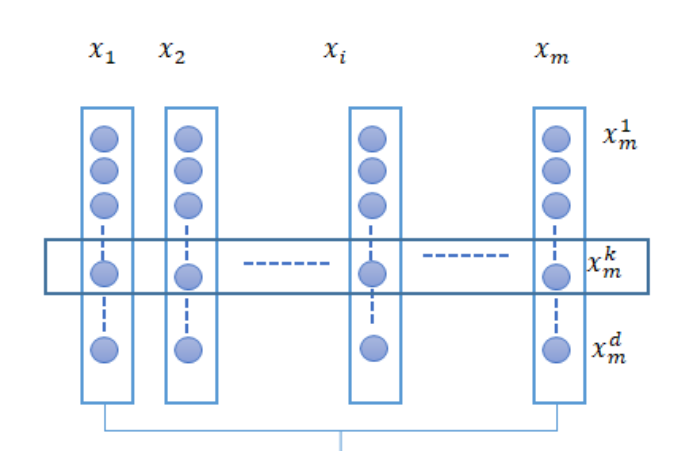}	
	\caption{Schematic diagram of quaternion batch normalization.}
	\label{p:mini}
\end{figure}

Consider a $mini$-$batch$ $\mathcal{B}=\{\xq_{1}^{k}, \xq_{2}^{k},\ldots, \xq_{m}^{k}\}.$ Since normalization is applied to each activation independently, as shown in Figure $\ref{p:mini}$.

Let the normalized values be $\hat{\xq}_{1}^{k}, \ldots,  \hat{\xq}_{m}^{k}$ and its linear transformation be $\yq_{1}^{k}, \ldots,  \yq_{m}^{k}$. We refer to the following transformation to denote the batch normalization transformation.

$$\phi_{\gamma^{k}, \beta^{k}}(x^{k}): x_{i}^{k}\rightarrow y_{i}^{k}, i=1, 2, \ldots. m $$

The batch normalized transformation applied to $mini$-$batch$ \xq~is shown in Algorithm $\ref{alg:1}$. In the algorithm, $\epsilon$ is a constant that is added to the $mini$-$batch$ variance for numerical stability.

\begin{algorithm}[H]
	\caption{Batch normalization transformations on mini-batch with respect to \xq}
	\label{alg:1}
	\begin{algorithmic}[1]
		\STATE{ \bf {Input}} :$mini$-$batch$~$\mathcal{B}=\{\xq_{1}, \xq_{2},\ldots, \xq_{m}\}$,$\xq_{i} \in \mathbb{Q}^{d}$. learning parameters ~$\gamma^{k},~\boldsymbol{\beta}^{k}, k=1, 2, \ldots,  d$.
		\STATE {\bf {Output}}:Batch normalization transformations $\phi_{\gamma^{k}, \boldsymbol{\beta}^{k}}(\xq^{k}): \xq_{i}^{k}\rightarrow \yq_{i}^{k},~ i=1, 2, \ldots,  m $.
		\STATE  $ \boldsymbol{\mu}^{k}_{\mathcal{B}}\leftarrow\frac{1}{m}\sum_{i=1}^{m}\xq^{k}_{i}$.
		\STATE $(\sigma^{k}_{\mathcal{B}})^{2} \leftarrow \frac{1}{m}\sum \limits_{i=1}^{m}\| \xq^{k}_{i}-\boldsymbol{\mu}^{k}_{\mathcal{B}}\|^{2}$.
		\STATE  $\hat{\xq}^{k}_{i}\leftarrow \frac{\xq^{k}_{i}-\boldsymbol{\mu}^{k}_{\mathcal{B}}} {\sqrt{(\sigma^{k}_{\mathcal{B}})^{2}+\epsilon}} $.
		\STATE $\yq^{k}_{i}\leftarrow \gamma^{k} \hat{\xq}^{k}_{i}+\boldsymbol{\beta}^{k}=\phi_{\gamma^{k},\boldsymbol{\beta}^{k}}(\xq_{i}^{k})$.
	\end{algorithmic}
\end{algorithm}

QBN is also a differentiable transform, which stabilizes the inputs to each network layer.
Layers receive $\phi_{\gamma, \boldsymbol{\beta}}(x)$ that previously received $x$ as input to each network. Once the network has been trained, we can use overall statistics instead of a small batch of statistics. Thus the final output of QBN is completely determined by the input $x$. It's worth noting that we use unbiased variance estimation $Var[x]= \frac{m}{m-1}E_{\mathcal{B}}[\sigma_{\mathcal{B}}^{2}]$, where the expectation obtained by training batches of size m and $\sigma_{\mathcal{B}}^{2}$ is their sample variance. Algorithm $\ref{alg:2}$ summarizes the process of training a quaternion batch normalization network.

\begin{algorithm}
	\caption{Training batch normalization network}
	\label{alg:2}
	\begin{algorithmic}[1] 
		\STATE{ \bf {Input}} : Network N, trainable parameters $\Theta$, activation set $\{x^{k}\}_{k=1}^{K}.$
		\STATE {\bf {Output}}: The completed training network $N_{test}$ with added batch normalization layers.
		
		\STATE $N_{train}\leftarrow N$.
		\FOR{$i = 1 \cdots K$}
		\STATE transform $\xq^{k}$ with $\yq^{k}=\phi_{\gamma^{k},\beta^{k}}(\xq^{k})$ (i.e., Algorithm 1).
		\STATE Modify the input of each layer in $N_{train}$: $\xq^{k} \leftarrow \yq^{k}$.
		\ENDFOR
		\STATE Training $N_{train}$ to optimize the parameters $\Theta \bigcup \{ \gamma^{k},\beta^{k}\}_{i=1}^{i=K} $.
		\STATE Fixed $N_{train}$ parameter:$N_{test} \leftarrow N_{train}$.
		\FOR{$i = 1 \cdots K$}
		\STATE Process training batches of $\mathcal{B}$:
		$$ E[\xq^{k}]\leftarrow E[\mu_{\mathcal{B}}^{k}], Var[\xq^{k}]\leftarrow \frac{m}{m-1}E[(\sigma_{\mathcal{B}}^{k})^{2}]$$.
		\STATE Replace $\yq=\phi_{\gamma, \beta}(\xq)$ in $N_{test}$ with 
		$$\yq^{k} \leftarrow \yq^{k}=\frac{\gamma}{\sqrt{Var[\xq^{k}]+\epsilon}}\xq^{k}+(\boldsymbol{\beta}-\frac{\gamma E[\xq^{k}]} {\sqrt{Var[\xq^{k}]+\epsilon}})$$.
		\ENDFOR
	\end{algorithmic}

\end{algorithm}

\subsection{Quaternion Generative Adversarial Network }
In original GANs, discriminator is usually a binary classification model, which is used to discriminate between real and fake images. The generator can be trained to generate realistic images close to the training set with a small error. The goal of the discriminator is to help improve the quality of the generated results, so that the trained discriminator can be fooled by unrealistic images. Nevertheless, it is frequently observed that the discriminator undergoes rapid training and acquires the capacity to distinguish between true and fake images within a few epochs. This, in turn, poses challenges in creating gradients for the generators inside the model to acquire knowledge \cite{r3}. Therefore we propose quaternion generator model to match its ability with the discriminator model:
\begin{equation}\label{e:QGAN}
	\begin{split}
		&\mathop{\rm min}\limits_{G} \mathop{\rm max}\limits_{D}V(G,D) = \\
		&E_{x\sim p_{data}}[{\rm log}(D(x))]+E_{z\sim p_{z}}[{\rm log}(1-D(\Gq(\zq))].
	\end{split}
\end{equation}
Compared with the previous Equation \ref{e:GAN}, the main change in Equation \ref{e:QGAN} is that all elements of the generator are quaternions, which will pose a great challenge to model training. How to design a generator model is what we need to concern about.

The quaternion generator consists of a series of quaternion deconvolution. Unlike the normal generative model, quaternion generator uses stepwise deconvolution instead of pooling layers. A quaternion deconvolution layer is followed by a quaternion batch normalization and a quaternion activation function action. The structure is shown in Figure \ref{f:C-G}. It can be seen that the entire generator network consists of a series of network blocks.

\section{The QGAN Method for Color Image inpainting}
We utilize the Quaternion Generative Adversarial Network (QGAN) for the purpose of color picture inpainting. 


The method of color image inpainting is separated into two main components. Initially, the whole training dataset is used to train QGAN. Afterwards, the pre-trained network is used to generate the absent color information. Subsequently, we present the inpainting model.

\subsection{Quaternion context loss function}
In order to inpaint the large missing regions, the remaining uncorrupted data is utilized. Quaternion context loss is proposed to capture this useful information. For semantic loss, choose just the number of paradigms between the generated sample ${\bf G(z)}$ and the uncorrupted part of the corrupted image $y$. The quaternion semantic loss can be denoted 
\begin{equation}\nonumber
	L_{c}(z|y, M)=\|M \cdot G(z)-M\cdot y   \|_{1}.
\end{equation}

The binary mask, denoted by $M$, is a quaternion matrix that shares the same dimensions as the image. In cases where a pixel is absent in the image, the corresponding position in the $M$ matrix is assigned a value of 0. Conversely, if a pixel is present, the corresponding position in the $M$ matrix is assigned a value of 1. Consequently, the $M$ matrix is a quaternion matrix with three imaginary components set to 0, effectively rendering it a real matrix. The symbol $\cdot$ is used to represent the dot product of matrices. The portion of the image that is corrupted, denoted as $(1-M)\cdot y$, is not utilized in the present loss function.

However, the above loss treats each pixel equally, which is not ideal. Consider the case where the center block is missing: a significant proportion of the loss will come from pixel locations far from the void, such as the background behind the face. Therefore, in order to find the correct encoding, more attention should be paid to pixel locations close to the missing region. To achieve this goal, a context loss with weights is used. It is assumed that the importance of an undamaged pixel is positively related to the number of damaged pixels around it. The significance of a pixel point is quantified by a weighted term denoted as $W$:
$$
W_{ij}=\left\{
\begin{aligned}
	\sum \sum_{(p,q) \in N_{ij}}\frac{1-M_{pq}}{| N_{ij}|}, \ M_{ij}\neq 0, \\
	0, \ M_{ij}= 0,\\
\end{aligned}
\right.$$
where $(i, j)$ is the pixel index, $W_{i}$ is the weight of the pixel at position $(i, j)$, $ N_{ij}$ is the set of neighbors of pixel $(i,j)$, and $|N_{ij}|$ is the number of neighbors of pixel $(i, j)$. Similarly, $W$ is a quaternion matrix with all three imaginary parts 0, which is essentially a real matrix. To summarize, the weighted semantic loss is a real matrices.
The weighted semantic loss is defined as the 1-parameter of the difference between the recovered image and the undamaged part as follows.
\begin{equation}\nonumber
	L_{c}(z)=\|W \cdot ({\bf G(z)}- y)\|_{1}.
\end{equation}
Let the size of the quaternion matrix ${\bf G(z)}$ be $m \times n$, according to the formula $\eqref{e:23}$ the quaternion matrix $l_{1}$ the definition of the paradigm number $L_{c}(z)$ can be written as
$$L_{c}(z)=\sum \limits_{i=1}^{m}\sum \limits_{j=1}^{n}|w_{ij}({\bf G(z)}_{ij}-y_{ij})|. $$

It is necessary to discuss the partial derivatives of $L_{c}(z)$ with respect to ${\bf G(z)}_{ij}$:

Let ${\bf G(z)}_{ij}=G_{ij}^{1}i+G_{ij}^{2}j+G_{ij}^{3}k$, $y_{ij}=y_{ij}^{1}i+y_{ij}^{2}j+y_{ij}^{3}k$.
When $w_{ij}=1$, 
\begin{equation}\label{GAN}
	\begin{split}
		&|\wq_{ij}({\bf G(z)}_{ij}-\yq_{ij})| \\
		&=\sqrt{(\Gq_{ij}^{1}-\yq_{ij}^{1})^{2}+(\Gq_{ij}^{2}-\yq_{ij}^{2})^{2}+(\Gq_{ij}^{3}-\yq_{ij}^{3})^{2}},
	\end{split}
\end{equation}

The partial derivative of $L_{c}(z)$ with respect to ${\bf G(z)}_{ij}$ is
\begin{equation}\nonumber
	\begin{split}
		\frac{\partial L_{c}(z)}{\partial {\bf G(z)}_{ij}}
		&=\frac{\partial \sum \limits_{i=1}^{m}\sum \limits_{j=1}^{n}|w_{ij}({\bf G(z)}_{ij}-y_{ij})|}{\partial {\bf G(z)}_{ij}}\\
		&=\frac{\partial |({\bf G(z)}_{ij}-y_{ij})|}{\partial {\bf G(z)}_{ij}}\\
		&=\frac{({\bf G(z)}_{ij}-y_{ij})^{T}}{|({\bf G(z)}_{ij}-y_{ij})|}.
	\end{split}
\end{equation}
When $w_{ij}=0$, $|w_{ij}({\bf G(z)}_{ij}-y_{ij})|=0,$ the partial derivative of $L_{c}(z)$ with respect to ${\bf G(z)}_{ij}$ is also 0. This can be expressed as:  $\frac{\partial L_{c}(z)}{\partial {\bf G( z)}_{ij}}=0.$
This gives the partial derivative of $L_{c}(z)$ with respect to ${\bf G(z)}_{ij}$.

\subsection{Quaternion priori loss function}
In order to make the reconstructed image similar to the samples extracted from the training set, the same loss function as that of the GAN is used to deceive the discriminator, denote the quaternion prior Loss as $L_{p}$
\begin{equation}\label{e:49}
	L_{p}(z)=\lambda {\rm log}(1-D({\bf G(z)})).
\end{equation}
where $\lambda$ is a parameter used to balance the two losses. Without $L_{p}$, the mapping from $y$ to $z$ could converge to a perceptually implausible result. The output of $D({\bf G(z)})$ is a real number, and the derivative of $L_{p}$ with respect to $D({\bf G(z)})$ is easily obtained as the upper derivative in the real domain.

The generator activation functions of the inpainter model are chosen in the same way as the pre-trained model.

\subsection{Image Inpainting Model}
The loss function of the inpainting model is the sum of the quaternion semantic loss and quaternion priori loss

\begin{equation}\label{e:50}
	L=L_{c}(z|y, M)+L_{p}(z).
\end{equation}

The corrupted image can be mapped to the most similar $z$ in the latent representation space. Image inpainting is optimized using Adam and $z$ is updated using total loss , resulting in $z^{'}$
\begin{equation}\nonumber
	z^{'}={\rm arg} \mathop{\rm min}\limits_{z} (L_{c}(z|y, M)+L_{p}(z)).
\end{equation}

After finding $z^{'}$, a generator is used to map $z^{'}$ to the image space, obtaining ${\bf G(z^{'})}$. Then, combining the missing parts of the input image $y$ with the corresponding missing parts of ${\bf G(z^{'})}$, we can get the inpainting result

\begin{equation}\nonumber
	\hat{y}=M \cdot y+(1-M)\cdot {\bf G(z^{'})}.
\end{equation}

The entire color image inpainting process is shown in Algorithm \ref{alg:3}.
\begin{algorithm}
	\caption{Quaternion Generative Adversarial Network Algorithm for Color Image inpainting}
	\label{alg:3}
	\begin{algorithmic}[1] 
		\STATE Pre-train QGAN.
		\STATE Input the original color image $y$. Denote the number of iterations as iter. Select sample points $z$ from the known noise distribution $p_{z}(z)$.
		\FOR {$i=1,2,\cdots,iter$}
		\STATE Sample point $z$, after generator to produce image ${\bf G(z)}$.
		\STATE Compute the loss function defined by Equation $\eqref{e:50}$.
		\STATE Backpropagation for updating $z$.
		\ENDFOR
		\STATE The final sample point $z^{'}$, and its corresponding image ${\bf G(z^{'})}$ generated by the generator, are obtained after iterations of updates.
		\STATE Performs Poisson image fusion on the initial inpainted image ${\bf G(z^{'})}$ to obtain the final inpainted color image $\hat{y}$.
	\end{algorithmic}

\end{algorithm}

More details about this section, such as the analysis of QBN, derivation of the backpropagation of QGAN and the construction of the quaternion loss function for QGAN image inpainting can be found in supplementary material.

\begin{figure*}[htb]
	\centering
		\includegraphics[width=0.95\textwidth,height=0.5\textwidth]{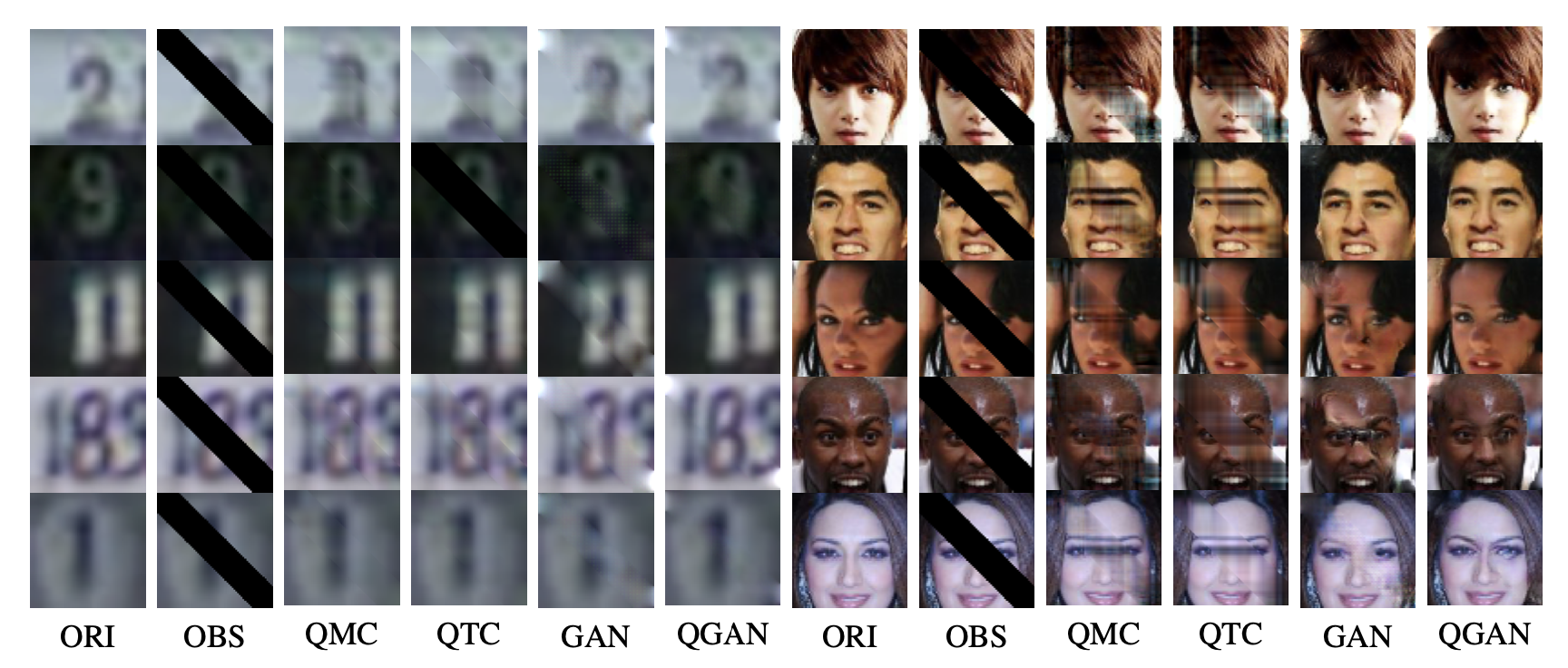}
	\caption{SVHN Database $36\%$ center Pixel Deletion Comparison of Various Inpainting Methods.}
	\label{f:diag}
\end{figure*}

\section{Experiments results}

Two databases to evaluate Quaternion Generative Adversarial Network Algorithms for Color Image inpainting : Street View House Number (SVHN)\cite{s31}\footnote{The SVHN database.  {http://ufldl.stanford.edu/housenumbers/}} and CelebA database (CelebA)\cite{s30}~\footnote{The CelebA database.  {http://mmlab.ie.cuhk.edu.hk/projects/CelebA.html}}.

It is worth noting that semantic inpainting does not attempt to reconstruct the underlying real image, the goal is to inpaint in the gaps with realistic content. Even a real image is one of many possibilities. The PSNR and SSIM values of the results of QGAN are compared to those of other algorithms.

$\bullet$ {\bf PSNR}: Peak Signal-to-Noise Ratio.  PSNR ranges from [0,100] and is inversely proportional to the image distortion.

$\bullet$ {\bf SSIM}: Structural Similarity.  SSIM can take values in the range [0,1], and is proportional to the image similarity.

\subsection{Pre-training}

Before proceeding with image inpainting, we need to pre-train the Generative Adversarial Neural Networks and Quaternion Generative Adversarial Neural Networks so that they gain the ability to inpaint the images.We just choose parts of the loss segments during the training process to display the process of training.

\begin{figure}[h]
	\centering
	\includegraphics[width=0.95\textwidth,height=0.6\textwidth]{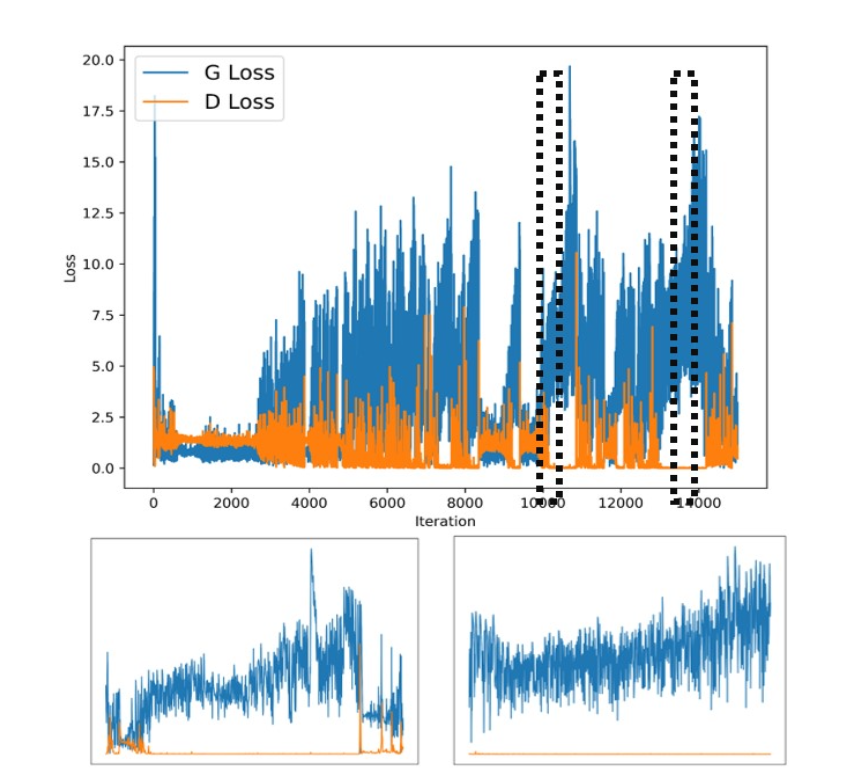}
	\caption{GAN loss map for 1-15000 iterations (with four loss map for iterations 10000-11000, 13000-14000).}
	\label{p:GAN_Loss}
	\vspace{-1.0em}
\end{figure}

From Figure \ref{p:GAN_Loss}, we can clearly see that both the loss values of the generator and the discriminator of the GAN are oscillating violently. In particular, in the two iteration regions boxed out in the main figure, the discriminator's loss value is infinitely close to 0, which is likely to damage the training results of the model and even lead to the failure of model training.

\begin{figure}[h]
	\centering
	\includegraphics[width=0.95\textwidth,height=0.6\textwidth]{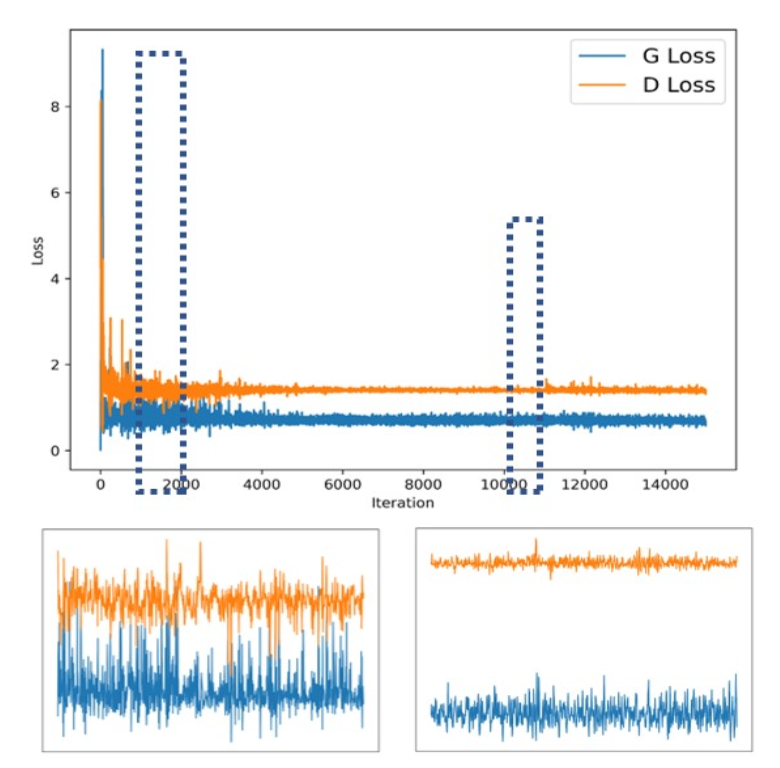}
	\caption{QGAN loss map for 1-15000 iterations (with two loss map for iterations 1000-2000, 10000-11000).}
	\label{p:QGAN_Loss}
	\vspace{-1.0em}
\end{figure}

In contrast, Figure \ref{p:QGAN_Loss} shows how the loss value of the generator and the loss value of the discriminator of QGAN change during the training process. We find that the two losses appear surprisingly stable after a certain number of iterations, with the generator loss hovering around 0.7 and the discriminator twofold. This phenomenon suggests that the introduction of quaternion deconvolution ensures stable training of the generative adversarial neural network's in the early stages of training, which guarantees that the generator acquires the same capabilities as the discriminator.

\subsection{QGAN Application to Color Image inpainting with Missing Central Block Pixels}

The QGAN algorithm is first used to inpaint color images with missing central blocks ($16\%$ pixels missing), which greatly increases the difficulty of the repair due to the excessive amount of missing continuous information in the central information.

\begin{table}[]
	\centering
	\caption{Comparison of PSNR and SSIM values of LRQMC, LRQTC, GAN and QGAN on database SVHN and CelebA with missing central block pixels}
	\label{tab:center}
	\centering
	\begin{tabular}{lllll}
		\hline
		Databases  & \multicolumn{2}{c}{SVHN}           & \multicolumn{2}{c}{CelebA}         \\ \cline{2-5} 
		Algorithms & PSNR             & SSIM            & PSNR             & SSIM            \\ \hline
		LRQMC \cite{s24}      & 25.2100          & 0.8880          & 20.2291          & 0.8440          \\
		LRQTC \cite{s16}      & {\ul 29.8151}    & 0.8905          & 22.1599          & 0.8491          \\
		GAN \cite{s15}       & 29.5473          & {\ul 0.9043}    & {\ul 27.9997}    & {\ul 0.8976}    \\
		Ours       & \textbf{34.8213} & \textbf{0.9625} & \textbf{30.6694} & \textbf{0.9322} \\ \hline
	\end{tabular}
\end{table}

From Table $\ref{tab:center}$, it can be seen that when inpainting in SVHN and CelebA databases, our results achieve optimal results on both PSNR and SSIM, with optimal values of 34.8213 and 0.9625. In comparison, LRQTC and LRQMC showed lower values during the inpainting process. This is because the LRQMC and LRQTC algorithms are not effective in inpainting this kind of images with missing central blocks, with obvious boundary traces in the visual effect, as shown in Figure $\ref{p:SVHN}$. And these two algorithms will fail to inpaint the image when the color of the image becomes darker. GAN algorithm will have a color mismatch when applied to image restoration. While our method not only takes into account the overall color distribution, but also the correlation of the three color channels, so the image inpainted by our method obviously improves the problems of the three algorithms mentioned above, and the image is much smoother and looks much more close to the original image.

\subsection{QGAN Application to color image inpainting with missing diag block pixels}

In this section we increase the difficulty of the restoration. It simulates a real-life situation where an image is diagonally damaged due to violent tearing. Here we apply the algorithm to inpaint color images with missing diagonal (36\% pixels missing from top left to bottom right). The method of comparison is the same as in the previous section.

\begin{table}[]
		\caption{Comparison of PSNR and SSIM values of LRQMC, LRQTC, GAN and QGAN on database SVHN and CelebA with missing diag block pixels}
		\label{tab:diag}
		\centering
	\begin{tabular}{lllll}
		\hline
		\multicolumn{1}{c}{Databases}  & \multicolumn{2}{c}{SVHN}                            & \multicolumn{2}{c}{CelebA}                          \\ \cline{2-5} 
		\multicolumn{1}{c}{Algorithms} & \multicolumn{1}{c}{PSNR} & \multicolumn{1}{c}{SSIM} & \multicolumn{1}{c}{PSNR} & \multicolumn{1}{c}{SSIM} \\ \hline
		LRQMC \cite{s24}                         & {\ul 25.8374}            & 0.7811                   & 18.7179                  & 0.6786                   \\
		LRQTC \cite{s16}                         & 24.9812                  & 0.7662                   & 19.6786                  & 0.6786                   \\
		GAN \cite{s15}                           & 23.7479                  & {\ul 0.8281}             & {\ul 21.6105}            & {\ul 0.8152}             \\
		Ours                           & \textbf{25.9241}         & \textbf{0.9197}          & \textbf{23.0758}         & \textbf{0.8538}          \\ \hline
	\end{tabular}

\end{table}

From Table $\ref{tab:diag}$, we can see that LRQMC, LRQTC and QGAN achieve good results in terms of PSNR values on the SVHN dataset. However, our algorithms remain optimal.
When we inpaint the CelebA dataset, the PSNR and SSIM values of all four algorithms show a decrease. This is because
that more details of faces need to be taken care of. The two algorithms LRQMC, LRQTC showed the fastest decrease. Combined with the partially repaired images shown in Figuer \ref{f:diag}. We can believe that the two algorithms LRQMC, LRQTC need to combine the information around the missing part to inpaint. However, this strategy is not applicable to the face inpainting task. Our algorithm inpaint after learning a large amount of face data, so it has a good recognition of the missing parts of the face dataset. Thus, our method can inpaint face images well. We can see that GAN will have color inpainting error. This is because the algorithm does not consider the correlation of the three channels. Whereas our algorithm represents each pixel as a pure imaginary quaternion for the restoration task, which considers the overall correlation.Therefore, it avoids the case of color error in the restored image and improves the restoration performance.

\subsection{Stability testing of more inpainting images}
The experiments in this section will test the stability of the QGAN algorithm in color image inpainting. We will randomly select 64 images at once from the test database for masking as observed images. Then, we calculated and presented the statistical values of PSNR and SSIM for the inpainting images in Table $\ref{tab:Statistical Values}$.

More details of the experiment will be shown in the supplementary material.

$\bullet$ {\bf Missing Centeral Block Pixels:} Here 64 images from the SVHN test database are randomly selected and given a missing central block (16\% pixels missing). 

$\bullet$ {\bf Missing Diag Block Pixels:} Here 64 images from the SVHN test database are randomly selected and given a missing diag block (36\% pixels missing).
 
\begin{table}[]
	\caption{Statistical Values of PSNR and SSIM for Inpainted Images}
	\label{tab:Statistical Values}
	\centering
	\begin{tabular}{ccccc}
		\hline
		\multicolumn{5}{c}{SVHN Database}                                          \\ \hline
		Missing Blocks     & \multicolumn{2}{c}{Center} & \multicolumn{2}{c}{Diag} \\
		Statistical Value  & PSNR         & SSIM        & PSNR        & SSIM       \\ \hline
		Maximun            & 46.770       & 0.992       & 32.093      & 0.960      \\
		Minimun            & 20.138       & 0.848       & 15.726      & 0.640      \\
		Average            & 32.609       & 0.937       & 24.952      & 0.880      \\
		Standard Deviation & 6.226        & 0.043       & 3.252       & 0.066      \\ \hline
		\multicolumn{5}{c}{CelebA Database}                                        \\ \hline
		Missing Blocks     & \multicolumn{2}{c}{Center} & \multicolumn{2}{c}{Diag} \\
		Statistical Value  & PSNR         & SSIM        & PSNR        & SSIM       \\ \hline
		Maximun            & 34.286       & 0.970       & 31.375      & 0.930      \\
		Minimun            & 21.985       & 0.853       & 16.649      & 0.757      \\
		Average            & 29.262       & 0.925       & 23.225      & 0.851      \\
		Standard Deviation & 3.063        & 0.028       & 2.727       & 0.038      \\ \hline
	\end{tabular}
\end{table}

As we have already emphasized, semantic inpainting is not view reconstruction of the underlying real image, but rather inpainting in the block with realistic content. Based on the above two tests we can see that QGAN has good inpainting on missing block. It is worth noting that when inpaint the images of CelebA dataset, the lowest value of SSIM is 0.757. Observing the corresponding images, we can see that the inpainted images already have the corresponding face shapes, but their color matches need to be further optimized, which is the reason for the low value of SSIM. We believe that increasing the number of iterations in Algorithm \ref{alg:3} can improve this situation. Most of the inpainted images closely approximate the real images. And the average values of these inpainting images are similar compared to the previous experiments. Therefore, in general, we can believe that the QGAN algorithm has superior inpainting ability and stability.

\section{Conclusion }

In this paper, a quaternion generative adversarial neural network model and its applications of inpainting color images with large area missing are investigated. To begin with, the constituents of QGAN are provided. The definition of quaternion deconvolution is provided by utilising the established features of quaternion convolution and its gradient back-propagation, along with quaternion rotation. And the quaternion batch normalization is added to stabilise the input distribution of the network layer. Then, the overall organizational framework of QGAN is given. The generator primarily comprises quaternion deconvolution and quaternion batch normalisation techniques, which effectively retain the correlation among the three channels. The experimental results show that the QGAN-based colour image semantic inpainting algorithm outperforms the existing algorithms in inpainting color images with large area missing.


\end{document}